%% file: preprint.tex
\documentclass{article}
\usepackage[preprint]{colm2025_conference}

\usepackage{microtype}
\usepackage[utf8]{inputenc} 
\usepackage[T1]{fontenc}    
\usepackage{hyperref}       
\usepackage{url}            

\usepackage{lineno}
\usepackage{adjustbox}
\usepackage{booktabs}       
\usepackage{amsfonts}       
\usepackage{nicefrac}       
\usepackage{amsmath}
\usepackage{marvosym}
\usepackage{graphicx}
\usepackage{colortbl}
\usepackage{subcaption}
\usepackage{changes}
\usepackage{listings}
\usepackage{mdframed}
\usepackage{caption}
\usepackage[most]{tcolorbox}
\usepackage{enumitem}
\usepackage{geometry}

\usepackage{hyperref}
\usepackage{url}
\usepackage[capitalise]{cleveref}
\usepackage{algorithm}
\usepackage{algpseudocode}

\usepackage{booktabs}
\usepackage{multirow}
\usepackage[table]{xcolor}

\usepackage{wrapfig}

\usepackage{longtable}
\usepackage{appendix}
\usepackage{titletoc} 

\usepackage{float} 
\usepackage{cite}

\renewcommand{\runninghead}{
  \Large\textbf{Knowledge\textcolor{blue}{\raisebox{-0.2\height}{\includegraphics[height=0.7cm]{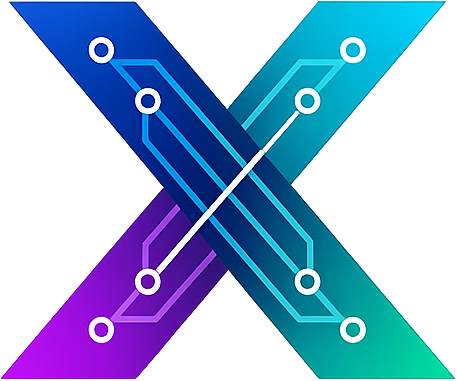}}} Lab}
}

\title{RE-Searcher: Robust Agentic Search with Goal-oriented Planning and Self-reflection}

\author{%
  Daocheng Fu$^{1, 2,\dag}$, Jianbiao Mei$^{3,2,\dag}$, Licheng Wen$^{2,4,5}$, Xuemeng Yang$^{2}$, Cheng Yang$^{2,6}$ \\
  \bf Rong Wu$^{3,2}$, Tao Hu$^{2}$, Siqi Li$^{3,2}$, Yufan Shen$^{2}$, Xinyu Cai$^{2}$, Pinlong Cai$^{2}$, Botian Shi$^{2,\textrm{\Letter}}$, \\ 
  \bf Yong Liu$^{3,\textrm{\Letter}}$,  Yu Qiao$^{2}$, \\ [1mm]
\textsuperscript{\rm 1} Fudan University,~~
\textsuperscript{\rm 2} Shanghai Artificial Intelligence Laboratory,
\textsuperscript{\rm 3} Zhejiang University \\
\textsuperscript{\rm 4} Shanghai Innovation Institute, 
\textsuperscript{\rm 5} Shanghai Jiao Tong University,
\textsuperscript{\rm 6} Central South University
\\ [1.5mm]
}

\definecolor{TagBlue}{RGB}{33,88,175}
\definecolor{TagPurple}{RGB}{128,0,128}
\definecolor{TagGreen}{RGB}{0,128,0}
\definecolor{TagOrange}{RGB}{230,120,0}
\newcommand{\opentag}[2]{\textcolor{#1}{\texttt{<#2>}}}
\newcommand{\closetag}[2]{\textcolor{#1}{\texttt{</#2>}}}
\newcommand{\think}{\opentag{TagOrange}{think}}
\newcommand{\thinkc}{\closetag{TagOrange}{think}}
\newcommand{\search}{\opentag{TagBlue}{search}}
\newcommand{\searchc}{\closetag{TagBlue}{search}}
\newcommand{\query}{\opentag{TagBlue}{query}}
\newcommand{\queryc}{\closetag{TagBlue}{query}}
\newcommand{\goal}{\opentag{TagBlue}{goal}}
\newcommand{\goalc}{\closetag{TagBlue}{goal}}
\newcommand{\reflect}{\opentag{TagPurple}{reflect}}
\newcommand{\reflectc}{\closetag{TagPurple}{reflect}}
\newcommand{\answer}{\opentag{TagGreen}{answer}}
\newcommand{\answerc}{\closetag{TagGreen}{answer}}

\newcommand{\method}{RE-Searcher}

\newtcolorbox{AIbox}[2][]{aibox, title=#2,#1}

\begin{document}

\maketitle

{
\renewcommand{\thefootnote}{}
\footnotetext{$^{\dag}$ Equal contribution, $^{\textrm{\Letter}}$ Corresponding authors.}
}

\input{sections/0_abstract}
\input{sections/1_introduction}

\input{sections/2_related_works}
\input{sections/3_pre_analysis}

\input{sections/4_methodology}

\input{sections/5_experiments}
\input{sections/6_conclusion}


\bibliographystyle{plain}
\bibliography{main}

\input{sections/appendix}

\end{document}

%% file: sections/0_abstract.tex
\begin{abstract}
Large language models (LLMs) excel at knowledge-intensive question answering and reasoning, yet their real-world deployment remains constrained by knowledge cutoff, hallucination, and limited interaction modalities. Augmenting LLMs with external search tools helps alleviate these issues, but it also exposes agents to a complex search environment in which small, plausible variations in query formulation can steer reasoning into unproductive trajectories and amplify errors. We present a systematic analysis that quantifies how environmental complexity induces fragile search behaviors and, in turn, degrades overall performance. To address this challenge, we propose a simple yet effective approach to instantiate a search agent, \method. During search, \method{} explicitly articulates a concrete search goal and subsequently reflects on whether the retrieved evidence satisfies that goal. This combination of goal-oriented planning and self-reflection enables \method{} to resist spurious cues in complex search environments and perform robust search. Extensive experiments show that our method improves search accuracy and achieves state-of-the-art results. Perturbation studies further demonstrate substantial resilience to noisy or misleading external signals, mitigating the fragility of the search process. We believe these findings offer practical guidance for integrating LLM-powered agents into more complex interactive environments and enabling more autonomous decision-making.
\end{abstract}

%% file: sections/1_introduction.tex
\section{Introduction}

Large language models (LLMs) have demonstrated remarkable performance in knowledge-intensive question answering and logical reasoning tasks~\cite{shao2024deepseekmath, li2025codei, minaee2024large}, and have gradually been deployed in real-world applications. Nevertheless, their further development remains constrained by several limitations: 
(1) \textbf{Knowledge cutoff}: model knowledge is restricted to the static pre-training corpus and cannot be updated in real time~\cite{shah2025beyond, cheng2024dated};  
(2) \textbf{Hallucination}: as probabilistic generators, LLMs inevitably produce content that is inconsistent with factual knowledge or user intent~\cite{ji2023towards,huang2025survey,tonmoy2024comprehensive};  
(3) \textbf{Interaction constraint}: models typically interact in a conversational form, restricting their capacity to perform more complex tasks~\cite{schick2023toolformer, yao2023react}.  
These challenges substantially limit the applicability of LLMs in open and dynamic real-world scenarios.  

\begin{figure}[t]
    \centering
    \includegraphics[width=0.99\linewidth]{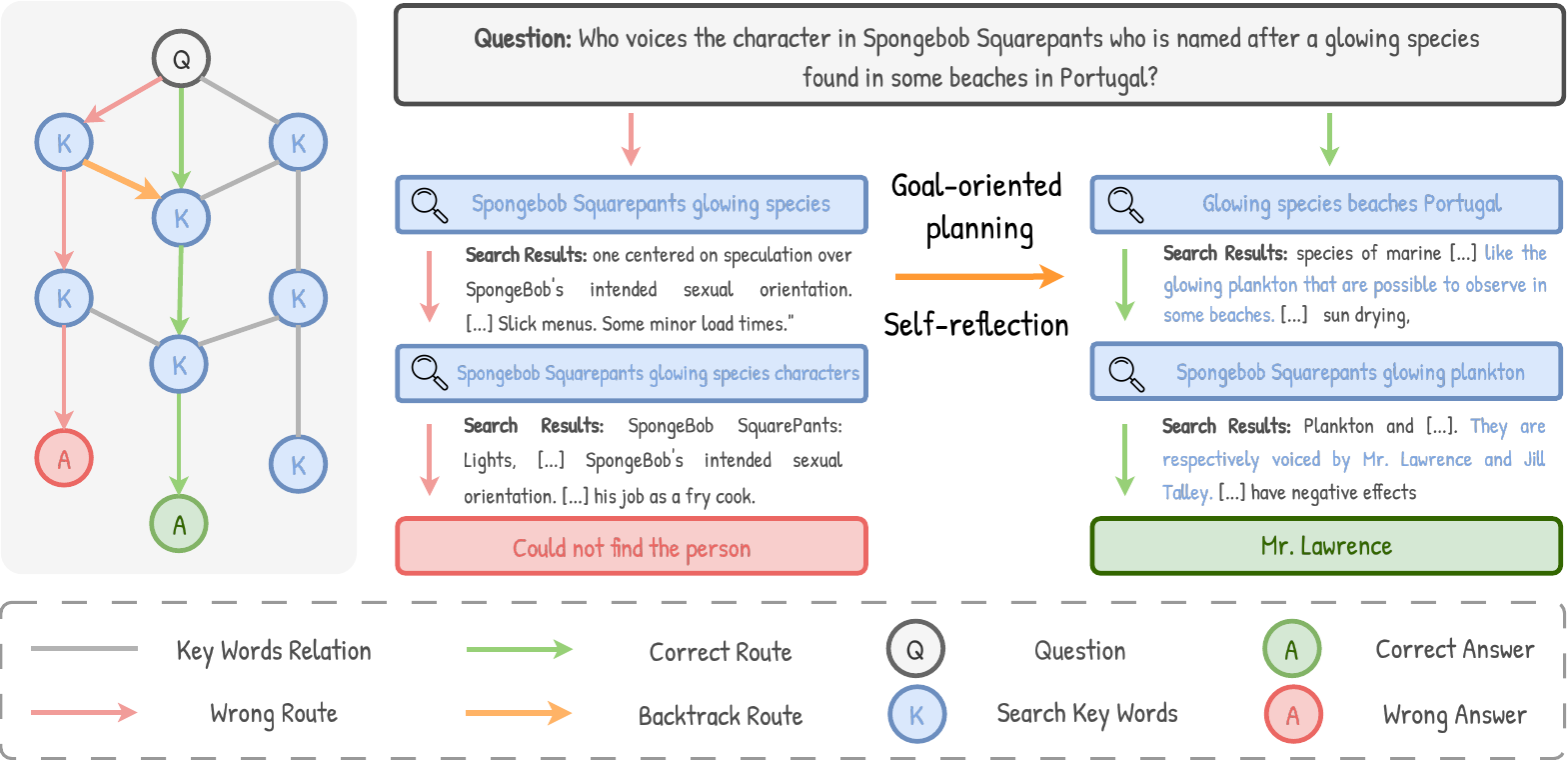}
    \caption{A search path can be viewed as a sample from the keyword graph. When receiving the same query, the search agent generates two distinct sets of keywords during two independent experiments. Although both sets of keywords are semantically sound, the retrieved results differed dramatically. Our \method, a search agent endowed with \textbf{goal-oriented planning} and \textbf{self-reflection} (orange arrow), can recover from such missteps and return to the correct trajectory, thereby enabling robust search behavior.}
    \vspace{-1.5em}
    \label{fig:problem_illustration}
\end{figure}

Recent research has sought to overcome these limitations by augmenting LLMs with external search tools, thereby constructing \emph{search agents}~\cite{jin2025search, zheng2025deepresearcher, wang2025stepsearch, hao2025dynasearcher}. By leveraging retrieval during response generation, such agents can extend the knowledge boundary of LLMs, alleviate hallucination, and enable more diverse downstream applications. However, while the search environment can enrich the information accessible to models, they can also introduce misleading evidence, resulting in degraded or erroneous response. In fact, as shown in \cref{sec:preliminary_analysis}, our preliminary analysis shows that the complexity of the search environment can lead to fragile interactions, which in turn amplify model errors and ultimately diminish task performance.
A simple illustrative case is presented in \cref{fig:problem_illustration}. When presented with the same query, the search agent issued two different sets of search keywords across two independent trials. Although both keyword choices were semantically reasonable, the retrieved results diverged dramatically. The erroneous trajectory (left) failed to yield useful information, and subsequent refinements along this trajectory could not recover the correct answer. By contrast, the correct trajectory (right) quickly retrieved the keyword ``plankton'' enabling the agent to find the correct answer in the second search step. 

Such variability and fragility of the search process pose considerable challenges for deploying LLMs in realistic settings. In contrast, humans are remarkably robust when operating under uncertain and dynamic conditions. Prior to executing a task, humans typically form explicit expectations of the desired outcome; after completion, they engage in reflection, evaluating whether the result meets expectations before deciding on subsequent actions. This process of \textbf{goal-oriented planning} and \textbf{self-reflection} enables humans to adapt flexibly to environmental complexity.

Inspired by this cognitive paradigm, we build a search agent, \textbf{\method}, that integrates goal-oriented planning with self-reflection. Specifically, in the search process, the agent is required to explicitly articulate its search goal and subsequently reflect on the quality of retrieved results. Our experiments demonstrate that this approach not only achieves state-of-the-art (SOTA) performance in search tasks but also substantially improves robustness. Further perturbation experiments reveal that our method enhances resilience to noisy or misleading external signals, thereby offering stronger adaptability to real-world, dynamic environments. Our contributions are listed below:

\begin{itemize}[leftmargin=2em, itemsep=0pt, topsep=2pt]
    \item We present a systematic analysis and quantification of how environmental complexity affects agent performance, underscoring the necessity of robustness for reliable deployment.
    \item We introduce a novel search agent, \textbf{{\method}}, that combines goal-oriented planning with self-reflection to mitigate the impact of noisy search results and correct potentially biased trajectories, showcasing a simple yet effective approach to achieving robust search performance.
    \item Extensive experiments demonstrate that {\method} improves search accuracy and robustness; perturbation analyses further validate the significant gains in resilience against external noise.
\end{itemize}

%% file: sections/2_related_works.tex
\section{Related Works}

Integrating external data is a pivotal strategy for overcoming the inherent limitations of Large Language Models (LLMs), notably knowledge cutoff and hallucination. The prevailing approaches can be broadly categorized into two paradigms: passive Retrieval-Augmented Generation (RAG) and proactive agentic search.

\subsection{Retrieval-augmented Generation}

Traditional RAG frameworks enhance model outputs by retrieving relevant information from an external corpus. This is typically achieved by encoding queries and knowledge passages into a shared vector space and fetching the nearest neighbors to augment the generation process for complex tasks~\cite{ma2023query, arslan2024survey, wu2025kg}. A significant drawback of these methods is their reliance on static, manually engineered prompts and workflows. Recent efforts have sought to improve RAG along two primary axes. On the retrieval front, methodologies like LightRAG~\cite{guo2024lightrag} and GraphRAG~\cite{edge2024local} leverage knowledge graphs to structure external data, facilitating more precise and contextually relevant information retrieval. On the generation front, works such as IRCoT~\cite{trivedi2022interleaving} integrate Chain-of-Thought (CoT) reasoning to refine both information seeking and synthesis. Meanwhile, AirRAG~\cite{feng2025airrag} employs Monte Carlo Tree Search (MCTS) to systematically explore diverse information pathways. Despite these advancements, these models remain fundamentally reactive; they do not proactively strategize on query formulation or dynamically adapt their reasoning based on retrieved results.

\subsection{Agentic Search-augmented Models}

A recent surge of interest has focused on developing autonomous agents that treat search engines as callable tools to support sophisticated reasoning. This agentic search paradigm for question-answering (QA) places a high demand on a model's planning and reasoning faculties, leading many researchers to turn to reinforcement learning (RL) for training. For instance, a series of works including Search-R1~\cite{jin2025search}, DeepResearcher~\cite{zheng2025deepresearcher}, and R1-Searcher++~\cite{song2025r1} have successfully applied RL algorithms like GRPO to train agents for multi-hop QA~\cite{yang2018hotpotqa, kwiatkowski2019natural}, significantly boosting their search and inference performance. StepSearch~\cite{wang2025stepsearch} refines this approach by introducing step-wise reward signals within a PPO framework, incentivizing productive actions at each stage of the search. Concurrently, DynaSearcher~\cite{hao2025dynasearcher} pioneers a dynamic knowledge graph that evolves during the search to guide exploration, while also leveraging heterogeneous data sources to enrich the agent's knowledge base. These contributions have substantially propelled the field forward, enabling models to more adeptly harness external knowledge for reasoning.

In this work, we build upon these foundations by performing a rigorous analysis of the search fragility brought by the complex search environment. We introduce a novel search agent designed to foster greater robustness during information retrieval, thereby elevating the quality and reliability of the model's final responses.

%% file: sections/3_pre_analysis.tex
\section{Preliminary Analysis}
\label{sec:preliminary_analysis}

The practical application of search agents is severely hampered by a significant instability in their outputs for search and question-answering. In this section, we begin by quantifying this stochasticity, and then leverage our findings to propose a simple but effective methodology aimed at enhancing the agents' overall performance and robustness.\footnote{We present the main results and our analysis here. Full experimental details are available in~\cref{sec:preliminary_exp}.}

\subsection{Stochasticity of Search Agent's Outputs}
\label{sec:randomness_of_output}

\begin{wrapfigure}{r}{0.45\textwidth}
    \vspace{-2em} 
    \centering
    \includegraphics[width=\linewidth]{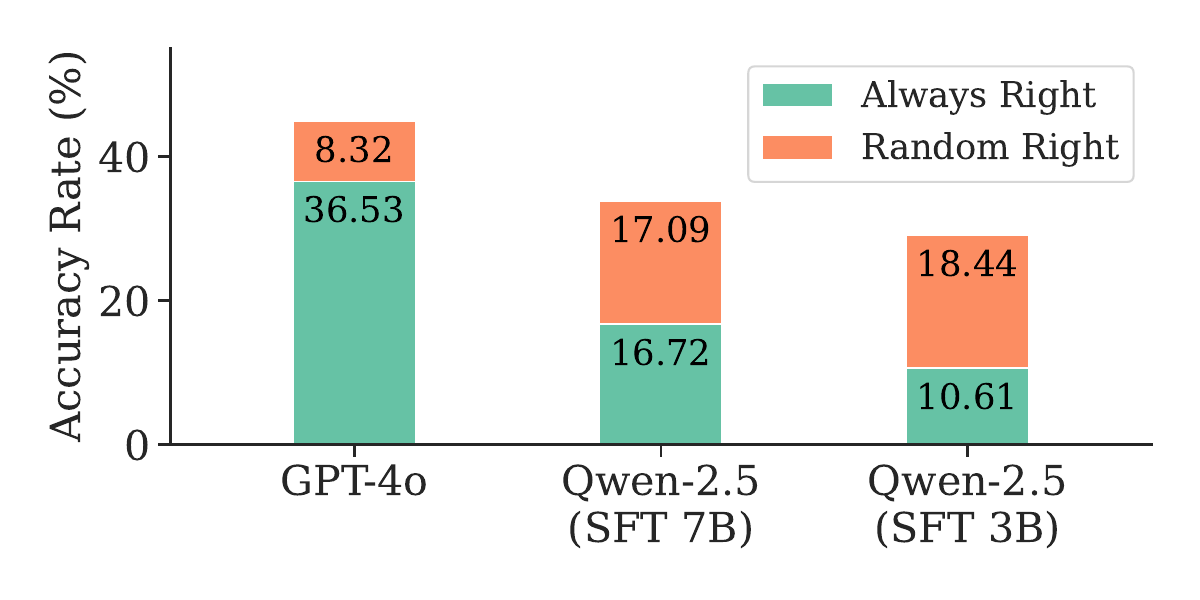}
    \vspace{-2em}
    \caption{Accuracy rate of search agents based on different models. \emph{always right} is the fraction of instances where all attempts are correct; \emph{random right} is the fraction where at least one attempt is correct}
    \label{fig:rollout_illustration}
    \vspace{-1.0em}
\end{wrapfigure}

To quantify the output instability, we evaluated search agents built upon various models. Each agent performed inference twice on an identical QA dataset. We classify questions as \textit{always right} if correctly answered in both runs, and as \textit{random right} if correct in only one. As illustrated in~\cref{fig:rollout_illustration}, GPT-4o~\cite{hurst2024gpt}, with its pre-trained tool-use capabilities~\cite{OpenAI_DeepResearch}, maintains a low, acceptable proportion of \textit{random right} outcomes. Conversely, Qwen2.5~\cite{qwen2025qwen25technicalreport}, which lacks this prior training, exhibits a \textit{random right} ratio that rivals or even surpasses its \textit{always right} ratio. This highlights a critical model instability that fundamentally limits the model's achievable performance.

\subsection{Fragility of the Search Process}
\label{sec:fragility_of_env}

\begin{wrapfigure}{r}{0.45\textwidth}
    \vspace{-2em} 
    \centering
    \includegraphics[width=\linewidth]{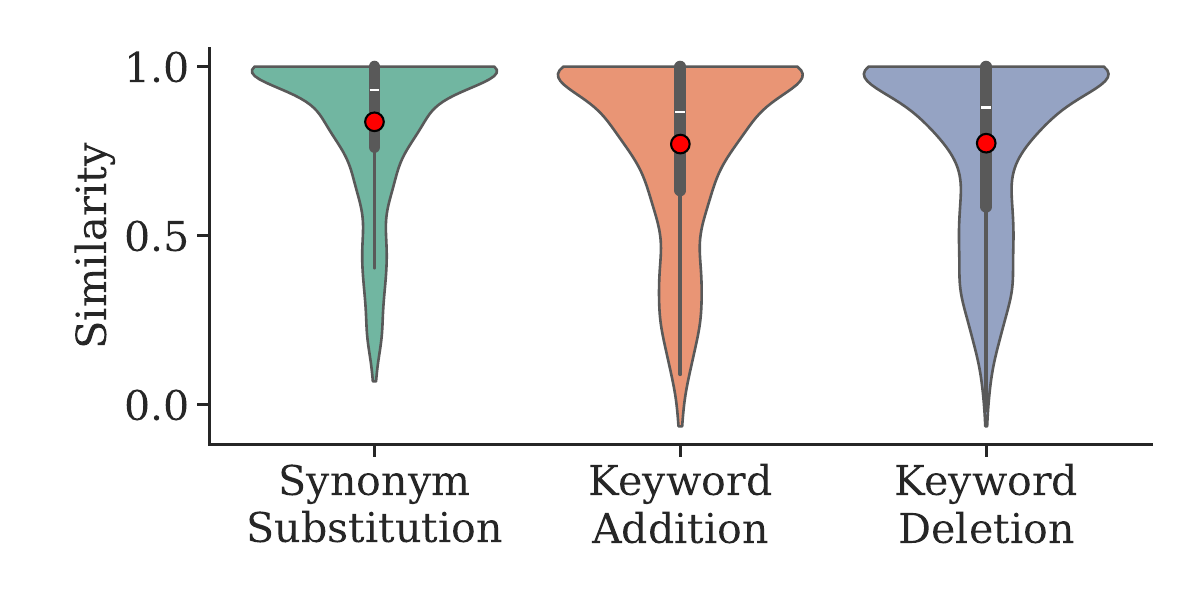}
    \vspace{-2em}
    \caption{Cosine similarity of the search results obtained from queries before and after perturbation; the red dot indicates the mean similarity.}
    \label{fig:perturbation_search_similarity}
    \vspace{-1.0em}
\end{wrapfigure}

Analyzing the search trajectories reveals a critical vulnerability: minuscule differences in search queries often lead to correct trajectories and incorrect ones. A single-word change in a query—such as a \textbf{synonym substitution}, \textbf{keyword addition}, or \textbf{keyword deletion}—can trigger drastically different results from the search engine. To demonstrate this, we applied these three types of micro-perturbations to search queries and measured the cosine similarity of the search results before and after. As shown in~\cref{fig:perturbation_search_similarity}, even these subtle changes frequently cause a sharp decline in semantic similarity, with many results dropping below a 0.6 threshold.

The complexity of search environment, therefore, acts as an amplifier for the agent's inherent stochasticity, often derailing its reasoning process towards erroneous conclusions. While a powerful model like GPT-4o can recover from such misleading signals, this underscores a general principle: an agent's ability to maintain a high-level goal and continuously self-reflect is paramount for robust performance. Motivated by this insight, our work focuses on explicitly training agents for \textbf{goal-oriented planning} and \textbf{self-reflection}. This equips them with the resilience needed to counteract the error amplification from the complex search environment.

\begin{figure}[t]
    \centering
    \includegraphics[width=0.99\linewidth]{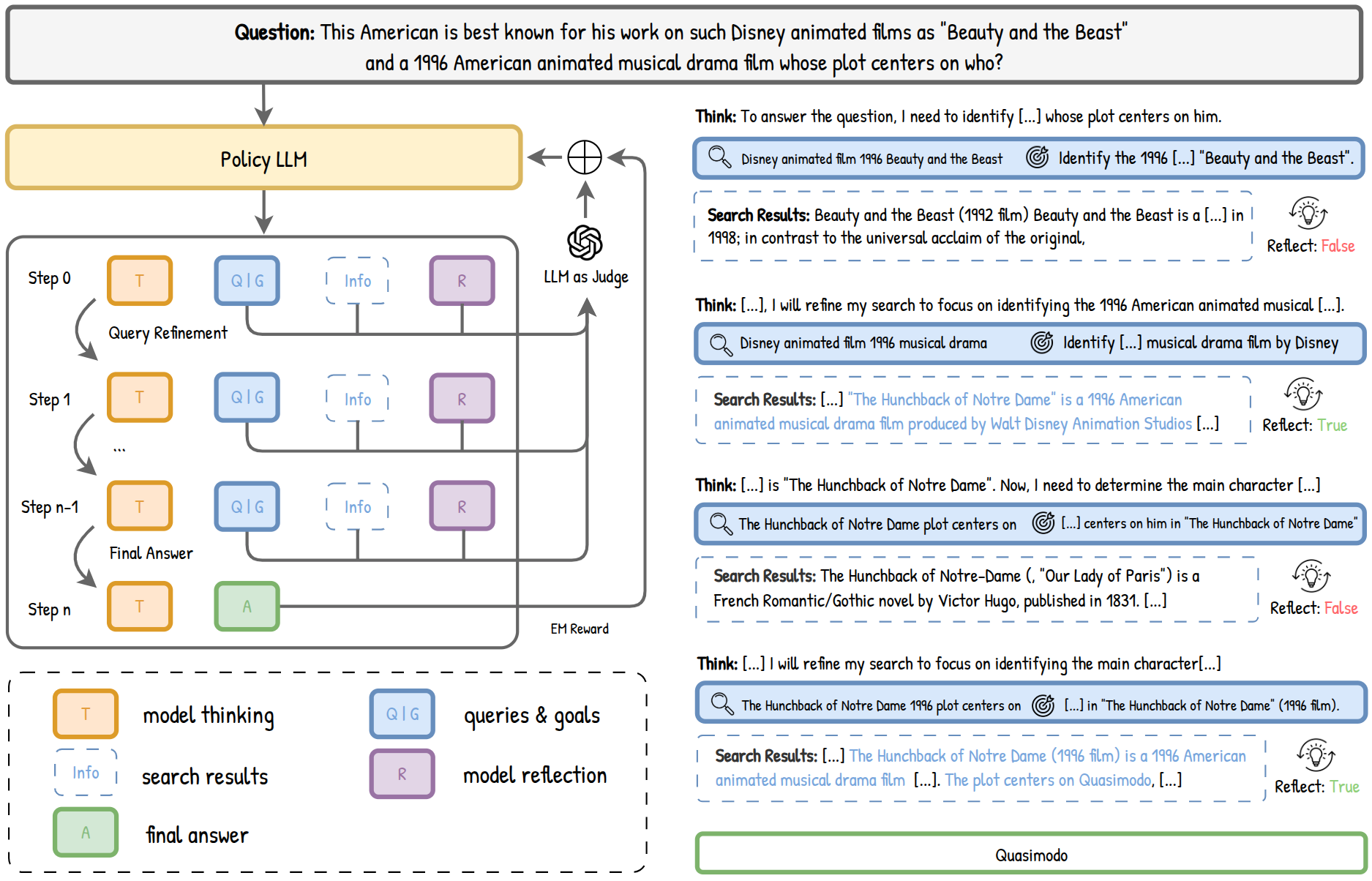}
    \caption{Illustration of the proposed training methods. Left: The model is required to explicitly plan its search goals during the search process and reflect on the results after obtaining them. An external LLM monitors the training model's reflection results to ensure that its judgments are correct. Right: The search trajectory made by the trained agentic model shows the correct reflection and goal planning.}
    \label{fig:framework}
    \vspace{-0.3em}
\end{figure}

%% file: sections/4_methodology.tex
\definecolor{color_think}{HTML}{F40103}
\definecolor{color_search}{HTML}{456195}
\definecolor{color_query}{HTML}{537CC8}
\definecolor{color_learnings}{HTML}{ED8137}
\definecolor{color_answer}{HTML}{759A5C}

\section{Methodology}

\input{tables/chat_template}

To enhance model robustness in complex search environments that often lead to fragile interactions, we aim to equip the agent with \textit{goal-oriented planning} and \textit{reflection} capabilities. As illustrated in~\cref{fig:framework}, during the training phase, the model is explicitly prompted to perform goal-oriented planning and reflection. Furthermore, an advanced LLM is employed to guide the model's reflective outputs. The resulting supervisory signal is then fed back to the primary model to refine its reflection accuracy.

\subsection{Explicit Searching with Reflection Behavior}
\label{sec:chat_template}

To enable the model to perform explicit search and reflection, we employ a structured generation template, as depicted in~\cref{tab:chat_template}, to constrain the model's output to one of three discrete actions at each turn: \textbf{Search}, \textbf{Reflect}, or \textbf{Answer}. Each action is preceded by a ``thought" process, where the model generates its rationale to ensure the subsequent output is coherent and well-founded.
The \textbf{Search} action is executed as follows: the model first analyzes the initial question and the \input{tables/algorithm1}
information gathered thus far to formulate a specific search \emph{goal} and a corresponding \emph{query}. A search engine then executes this \emph{`query'} and returns the results. 
During the \textbf{Reflect} action, the model evaluates whether the retrieved search results align with the stated \emph{goal}. If the goal is met, the model confirms this with a \emph{TRUE} judgment and proceeds to formulate a new search \emph{goal} and \emph{query}. 
Conversely, if the results are unsatisfactory, the model refines the \emph{query} and re-initiates the search process to fulfill the original goal.
Finally, once all necessary information has been gathered and all sub-goals are satisfied, the model transitions to the \textbf{Answer} action, synthesizing the collected evidence to produce the final response to the user's question. The full search process is shown in~\cref{alg:search_reflection}.

To ensure the model adheres to the required output format during training, we construct a small set of chain-of-thought (CoT) interaction trajectories (approximately 1K) as a warm-up. We build an LLM agent based on a strong instruction-following model (GPT-4o) to generate interactions that conform to the above protocol, including the thought process, search steps, reflection, and final answer. These data are then used to fine-tune the base model, enabling it to produce outputs in the desired format.

\subsection{GRPO with Search Engine}

The use of reinforcement learning algorithms to improve the search capabilities of models has been widely validated~\cite{li2025search, wang2025stepsearch, hao2025dynasearcher}. In this work, to mitigate the demand for computational resources, we employ Group Relative Policy Optimization (GRPO)~\cite{shao2024deepseekmath} to train the model's search and reflection abilities. For each input question $x$ in GRPO, a group of $G$ rollout trajectories, denoted as $\tau=\{y_i\}_{i=1}^{G}$, is generated using the preceding policy $\pi_{old}$, the current policy model $\pi_\theta$ is subsequently optimized by maximizing the objective function:

\begin{equation} \label{eq:grpo}
\mathcal{L}(\theta) = \mathbb{E}_{\substack{ x\sim \mathcal{D},\{y_i\}_{i=1}^G \\ y_i \sim \pi_{old}(\cdot|x)}} \left[ \frac{1}{G} \sum_{i=1}^{G} \min \left( r_i(\theta) {A}_i, \text{clip}(r_i(\theta), 1-\epsilon, 1+\epsilon) {A}_i \right) - \beta\mathbb{D}_{KL}[\pi_\theta||\pi_{ref}] \right]
\end{equation} 

where $\pi_{ref}$ denotes reference model, $r_i(\theta)=\frac{\pi_\theta(y_i|x)}{\pi_{old}(y_i|x)}$. $\epsilon$ and $\beta$ are hyperparameter. $A_i$ represents the advantage, computed based on the relative rewards (which will be mentioned in~\cref{sec:reward_design}) of outputs within each group. As mentioned in~\cref{sec:chat_template}, in each rollout, the model will take search actions using \search{}\searchc{} tags, and the retrieved tokens that are tagged by \texttt{<learnings></learnings>} will be masked when calculating the loss.

\subsection{Reflection Supervision Through LLM as Judge}
\label{sec:reward_design}
After the warm-up phase, the model has learned to output in the desired format to some extent. To further enforce the correct format during the reinforcement learning stage, we integrate format constraints with the factual reward. Specifically, the output trajectory is encouraged to continuously include the actions of search and reflection, with the final action being the answer. Following the method in \cite{jin2025search}, we combine the format reward with the factual reward as follows:

\begin{equation}
     r_{em\_format} =
\begin{cases}
1-0.2\cdot\mathrm{FM}(\tau_{pred}), & \text{if } \mathrm{EM}(a_{\mathrm{pred}}, a_{\mathrm{gt}}) = 1,
 \\
0.2\cdot\mathrm{FM}(\tau_{pred}), & \text{if } \mathrm{EM}(a_{\mathrm{pred}}, a_{\mathrm{gt}}) = 0.
\end{cases}
\end{equation} where $\mathrm{EM}$ is the exact match function and $\mathrm{FM}$ evaluates whether the predicted trajectory $\tau_{pred}$ follows the required output format. $a_{pred}$ and $a_{gt}$ denote the predicted and ground-truth answers, respectively.

We further employ model-based evaluation, i.e., an LLM as a judge to guide the model's reflection process. Specifically, we prompt GPT-4o-mini with a triple input, comprising the search goal, the search result, and the judgment, to evaluate whether the model's reflection judgment is correct. The reflection reward is weighted and added to the factual reward with format constraints for the final reward:
\begin{equation}
    r = r_{em\_format} + \sum_{i}0.1*\mathrm{MBE}(g_i, s_i, v_i)
\end{equation} where $\mathrm{MBE}$ denotes the model-based evaluation. $(g_i, s_i, v_i)$ is the search goal, the search result, and the judgment for the $i$-th search action.

\input{tables/deterministic}



%% file: tables/chat_template.tex
\begin{table}[b]
\centering
\caption{Chat Template for \method, when the model answers questions, it needs to think, plan, search, and reflect to ensure the robustness of the search path.}
\label{tab:chat_template}
\renewcommand{\arraystretch}{1.2}
    \resizebox{\linewidth}{!}{%
        \begin{tabular}{p{0.99\linewidth}} 
        \toprule
    As an expert researcher, provide precise answers to the given question. 
    When new information arrives, first reason within \think{} and \thinkc{} tags to analyze the question and determine search keywords. 
    Each search must include a clear \goal{} specifying the information you aim to find, along with \query{} items combining initial questions with collected information (e.g., \search{} \query{} QUERY \queryc{} \goal{} GOAL \goalc{} \searchc{}). 
    After receiving search results in \texttt{<learnings></learnings>} tags, reflect on whether they meet your goal using \think{} for analysis, then explicitly state the outcome in \reflect{} True/False \reflectc{} (True = goal met, False = needs refinement). 
    If knowledge gaps exist, perform up to five iterative searches with refined goals/queries. 
    When sufficient information is obtained, present the final answer within \answer{} \answerc{} tags. \\
        \bottomrule
        \end{tabular}%
    }
\end{table}

%% file: tables/algorithm1.tex
\begin{wrapfigure}{r}{0.55\textwidth}
\vspace{-0.9em}
\centering
\tiny
\resizebox{\linewidth}{!}{  
\begin{minipage}{\linewidth}
\begin{algorithm}[H]
    \small
    \caption{Iterative Search and Reflection}
    \label{alg:search_reflection}
        \begin{algorithmic}[1]
        \Require User question $Q$
        \State \textbf{Initialize:} Context $\mathcal{C} \gets \{Q\}$, $\mathcal{G}_{\text{pending}} \gets \emptyset$, $\mathcal{G}_{\text{completed}} \gets \emptyset$
        \State Generate an initial search goal based on the input question $Q$ and add it to $\mathcal{G}_{\text{pending}}$.
        
        \While{$\mathcal{G}_{\text{pending}} \neq \emptyset$}
            \State Get current goal $g_{\text{current}}$ from $\mathcal{G}_{\text{pending}}$
            \State $\text{is\_goal\_met} \gets \text{FALSE}$
            
            \While{NOT $\text{is\_goal\_met}$}
                \State Generate query $q$ based on $g_{\text{current}}$ and context $\mathcal{C}$.
                \State Retrieve results $R \gets \text{SearchEngine}(q)$.
                \State Update context: $\mathcal{C} \gets \mathcal{C} \cup \{R\}$.
                
                \State Generate judgment $J \gets \text{Reflect}(R, g_{\text{current}})$.
                
                \If{$J = \text{TRUE}$}
                    \State $\text{is\_goal\_met} \gets \text{TRUE}$
                    \State Move $g_{\text{current}}$ from $\mathcal{G}_{\text{pending}}$ to $\mathcal{G}_{\text{completed}}$.
                    \State Identify a new search goal ${g}_{\text{new}}$ based on $\mathcal{C}$.
                    \State $\mathcal{G}_{\text{pending}} \gets \mathcal{G}_{\text{pending}} \cup \{g_{\text{new}}\}$.
                \EndIf
            \EndWhile
        \EndWhile
        
        \State Generate final answer $A$ based on the complete context $\mathcal{C}$.
        \State \Return $A$
        \end{algorithmic}
\end{algorithm}
\end{minipage}
}
\vspace{-1.5em}
\end{wrapfigure}

%% file: tables/deterministic.tex
\begin{table}[t]
    \centering
    \small
    \setlength{\tabcolsep}{10pt}
    \renewcommand{\arraystretch}{1.05}
    \caption{Exact Match (EM) metrics on question-answering tasks. The best performance is set in \textbf{bold}. Our {\method} outperforms all baselines across most in/out-of-domain datasets using both Qwen2.5-3B-Instruct and Qwen2.5-7B-Instruct as base model.}
    \vspace{5pt}
    \definecolor{mycolor}{rgb}{0.64, 0.87, 0.93}
    \label{tab:deterministic}
    \scalebox{0.8}{
    \begin{tabular}{lcc|ccccc>{\columncolor{gray!10}}c}
        \toprule
        \multirow{2}{*}{\textbf{Methods}} & \multicolumn{2}{c}{\textbf{In domain}} & \multicolumn{5}{c}{\textbf{Out of domain}} & \multirow{2}{*}{\textbf{Avg.}}\\
        \cmidrule{2-8}
         & \textbf{NQ} & \textbf{HotpotQA} & \textbf{TriviaQA} & \textbf{PopQA} & \textbf{2wiki} & \textbf{Musique} & \textbf{Bamboogle} &  \\
        \midrule
        \multicolumn{8}{l}{\textbf{Qwen2.5-3B}} \\
        Direct Inference & 0.106 & 0.149 & 0.288 & 0.108 & 0.244 & 0.020 & 0.024 & 0.134 \\
        CoT & 0.023 & 0.021 & 0.032 & 0.005 & 0.021 & 0.002 & 0.000 & 0.015 \\
        IRCoT & 0.111 & 0.164 & 0.312 & 0.200 & 0.171 & 0.067 & 0.240 & 0.181 \\
        Search-o1 & 0.238 & 0.221 & 0.472 & 0.262 & 0.218 & 0.054 & 0.320 & 0.255 \\
        RAG & 0.348 & 0.255 & 0.544 & 0.387 & 0.226 & 0.047 & 0.080 & 0.270  \\
        SFT & 0.249 & 0.186 & 0.292 & 0.104 & 0.248 & 0.044 & 0.112 & 0.176  \\
        R1-base & 0.226 & 0.201 & 0.455 & 0.173 & 0.268 & 0.055 & 0.224 & 0.229  \\
        R1-instruct & 0.210 & 0.208 & 0.449 & 0.171 & 0.275 & 0.060 & 0.192 & 0.224  \\
        Search-R1-base & 0.406 & 0.284 & 0.587 & {0.435} & 0.273 & 0.049 & 0.088 & 0.303  \\
        Search-R1-instruct & 0.341 & 0.324 & 0.545 & 0.378 & 0.319 & 0.103 & 0.264 & 0.325 \\
        O$^2$-Searcher & \textbf{0.444} & {0.388} & {0.597} & 0.429 & {0.374} & {0.160} & {0.344} & {0.391} \\
        ZeroSearch-base & 0.430 & 0.338 & \textbf{0.616} & 0.414 & 0.346 & 0.130 &  0.139 & 0.345 \\
        ZeroSearch-instruct & 0.414 & 0.274 & 0.574 &  \textbf{0.448 }& 0.300 & 0.098 & 0.111 & 0.317 \\
        OTC & \textbf{0.444} & 0.365 & 0.608 & 0.441 & 0.341 & 0.124 & 0.266 & 0.370\\
        \rowcolor{mycolor!25} {\method} (ours) & 0.419 & \textbf{0.404}  & 0.600  & 0.416  & \textbf{0.420}  & \textbf{0.166}  & \textbf{0.408}  & \textbf{0.405} \\
        \midrule
        \multicolumn{8}{l}{\textbf{Qwen2.5-7B}} \\
        Direct Inference & 0.134 & 0.183 & 0.408 & 0.140 & 0.250 & 0.031 & 0.120 & 0.181 \\
        CoT & 0.048 & 0.092 & 0.185 & 0.054 & 0.111 & 0.022 & 0.232 & 0.106 \\
        IRCoT & 0.224 & 0.133 & 0.478 & 0.301 & 0.149 & 0.072 & 0.224 & 0.226 \\
        Search-o1 & 0.151 & 0.187 & 0.443 & 0.131 & 0.176 & 0.058 & 0.296 & 0.206 \\
        RAG & 0.349 & 0.299 & 0.585 & 0.392 & 0.235 & 0.058 & 0.208 & 0.304 \\
        SFT & 0.318 & 0.217 & 0.354 & 0.121 & 0.259 & 0.066 & 0.112 & 0.207  \\
        R1-base & 0.297 & 0.242 & 0.539 & 0.202 & 0.273 & 0.083 & 0.296 & 0.276  \\
        R1-instruct & 0.270 & 0.237 & 0.537 & 0.199 & 0.292 & 0.072 & 0.293 & 0.271  \\
        Search-R1-base & 0.480 & 0.433 & 0.638 & 0.457 & 0.382 & \textbf{0.196} & 0.432 & 0.431  \\
        Search-R1-instruct & 0.393 & 0.370 & 0.610 & 0.397 & 0.414 & 0.146 & 0.368 & 0.385 \\
        ZeroSearch-base & 0.424 & 0.320 & \textbf{0.664} & \textbf{0.604} & 0.340 & 0.180 & 0.333 & 0.409 \\
        ZeroSearch-instruct & 0.436 & 0.346 & 0.652 &  0.488 & 0.352 & 0.184 & 0.278 & 0.391 \\
        OTC & 0.444 & 0.366 & 0.597 & 0.431 & 0.311 & 0.130 & 0.250 & 0.361\\
        \rowcolor{mycolor!25} {\method} (ours) & \textbf{0.453}  & \textbf{0.437} & 0.638 & 0.454 & \textbf{0.473}  & 0.194 & \textbf{0.496} & \textbf{0.449}\\
        
        \bottomrule
    \end{tabular}}
    \vspace{-10pt}
\end{table}

%% file: sections/5_experiments.tex
\section{Experiments}

In this section, we design and conduct a series of experiments to answer the following key research questions (RQs):

\begin{itemize}[leftmargin=3em, itemsep=0pt, topsep=2pt]
    \item \textbf{RQ1}: Does the reflection-augmented framework improve problem-solving capabilities in search tasks? (\cref{exp:analysis})
    \item \textbf{RQ2}: To what extent does the reflection mechanism mitigate the negative impacts of search fragility? (\cref{exp:nim})
    \item \textbf{RQ3}: How much does the proposed framework enhance the model's robustness against external disturbances? (\cref{exp:disturb})
\end{itemize}

\subsection{Implementation Details} 
\label{exp:imp}

\textbf{Setup}. We adopt Qwen2.5-3B-Instruct and Qwen2.5-7B-Instruct \cite{yang2024qwen2} as the backbone models of our proposed {\method}. For the cold start stage, we utilize the Adam optimizer with an initial learning rate of $1 \times 10^{-5}$  and a warm-up ratio of 0.1. This stage is conducted on 8 A100 GPUs for 2 epochs.
During the RL training stage, we employ the Verl framework  \footnote{https://github.com/volcengine/verl}. We optimize the policy model using the GRPO algorithm. At each training step on 8 A100 GPUs, we sample a batch of 64 prompts, generating 8 rollout trajectories for each. The model is updated with the Adam optimizer at a learning rate of $1 \times 10^{-6}$. For GRPO, we set the KL divergence regularization coefficient $\beta$ to 0.001 and the clip ratio $\epsilon$ to 0.2. The maximum sequence length is configured to be 10$k$ tokens, while retrieved content is restricted to 2$k$ tokens, and the maximum number of action steps is 11. To accelerate LLM rollouts, we leverage vLLM \footnote{https://github.com/vllm-project/vllm}  with a tensor parallel size of 1 and a GPU memory utilization ratio of 0.85. For rollout sampling, we use a temperature of 1.0 and a top-$p$ value of 1.0.

\textbf{Datasets.} We assess our proposed \text{\method} on both in-domain and out-of-domain datasets. The models are trained on in-domain datasets, including NQ \cite{kwiatkowski2019natural} and HotpotQA \cite{yang2018hotpotqa}, while the out-of-domain datasets encompass TriviaQA \cite{joshi2017triviaqa}, PopQA \cite{mallen2022not}, 2WikiMultiHopQA \cite{ho2020constructing}, Musique \cite{trivedi2022musique}, and Bamboogle \cite{press2022measuring}. In total, these validation tests involve 51,953 questions with corresponding ground-truth answers.

\textbf{Baselines.} We follow the setting of Search-R1 \cite{jin2025search} and compare our {\method} against two categories of methods: (1) CoT-based approaches, including CoT \cite{wei2022chain}, RAG \cite{lewis2020retrieval}, IRCoT \cite{trivedi2022interleaving}, and Search-o1 \cite{li2025search}. These methods leverage Chain-of-Thought reasoning either for direct inference or in combination with Retrieval-Augmented Generation (RAG). (2) Train-based methods, such as Supervised Fine-Tuning (SFT) \cite{chung2024scaling}, DeepSeek-R1 \cite{guo2025deepseek}, Search-R1 \cite{jin2025search}, ZeroSearch \cite{sun2025zerosearch}, O$^2$-Searcher \cite{mei20252}, and OTC \cite{wang2025otc}. SFT and DeepSeek-R1 perform reasoning and answer steps without a search engine, whereas other methods incorporate a local search engine.

\textbf{Metrics.} The Exact Match (EM) and F1 metrics are applied, following \cite{yu2024rankrag, jin2025search}.

\input{tables/ab1}
\input{tables/ab2}

\vspace{-1em}

\subsection{Effectiveness of Self-Reflection Mechanism} 
\label{exp:analysis}

\subsubsection{Improvement of Searching Ability}

We conducted a comprehensive evaluation of {\method} on both in-domain and out-of-domain tasks, with detailed results presented in ~\cref{tab:deterministic}. The findings clearly indicate that our method establishes a new state-of-the-art, outperforming all baseline methods across both the 7B and 3B model scales. Using the Qwen2.5-7B-instruct model as the backbone, {\method} achieves the highest average EM score of 0.449, surpassing all other approaches. Notably, it secures top performance on both in-domain datasets, NQ and HotpotQA, demonstrating its proficiency on familiar tasks. Furthermore, it shows exceptional generalization to out-of-domain datasets, achieving the best scores on 2WikiMultiHopQA and Bamboogle. Compared to recent RL-based baselines, such as Search-R1 and ZeroSearch, our method provides a significant improvement in average performance, underscoring the effectiveness of our approach.
To validate the scalability and efficiency of our method, we also evaluated it on the smaller Qwen2.5-3B-instruct model. The results reinforce our claims, as {\method} again achieves the highest average EM score of 0.405, outperforming competitive methods like O$^2$-Searcher and OTC. This consistent superiority highlights the scalability and robust effectiveness of our approach.
\cref{fig:framework} shows a search trajectory of \method. The model plans the search goal for each search and reflects on whether the retrieved content meets the requirements. During the third search, the search engine incorrectly returned information about a novel with the same title. Through reflection, the model simply modified a single keyword and obtained the correct result.

\begin{wrapfigure}{r}{0.55\textwidth}
    \centering
    \vspace{-1em}
    \includegraphics[width=\linewidth]{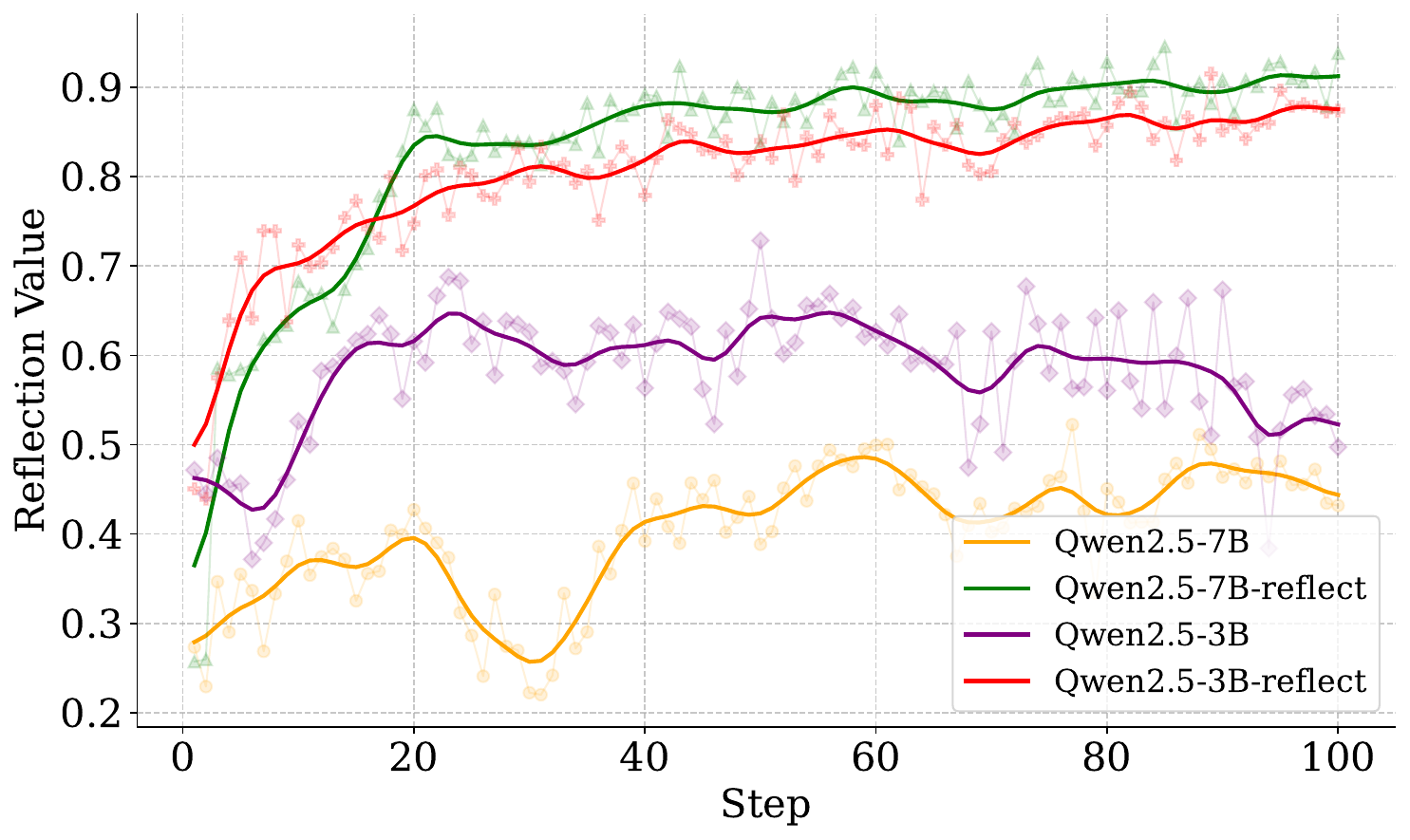}
    \vspace{-1.5em}
    \caption{The training dynamics of the reflection value of different models.}
    \label{fig:training_curve}
\end{wrapfigure}

\subsubsection{Analysis on reflection reward}

We analyze the impact of the reflection reward on training dynamics. As illustrated in~\cref{fig:training_curve}, the model trained without this reward exhibits a reflection score that hovers around 0.5. This indicates a near-random judgment on the consistency between the retrieved information and the search goal, underscoring the importance of the explicit guidance provided by the LLM-as-judge. In contrast, with the reflection reward, the score stabilizes at a higher value, demonstrating that the model learns a consistent and effective reflection policy.
These training dynamics are corroborated by quantitative results on the validation set. As shown in~\cref{tab:ab1} and  \cref{tab:ab2}, removing the reflection reward leads to a consistent performance drop across both in-domain and out-of-domain datasets, as well as all evaluated multi-hop datasets. Conversely, its inclusion yields significant improvements, particularly on the more challenging 2wiki (+0.062 in both EM and F1) and Bamboogle (+0.069 in EM and +0.045 in F1) datasets. While the gains on Musique are more modest, they remain consistently positive across both metrics.

\begin{figure}[t]
    \centering
    \includegraphics[width=0.80\linewidth]{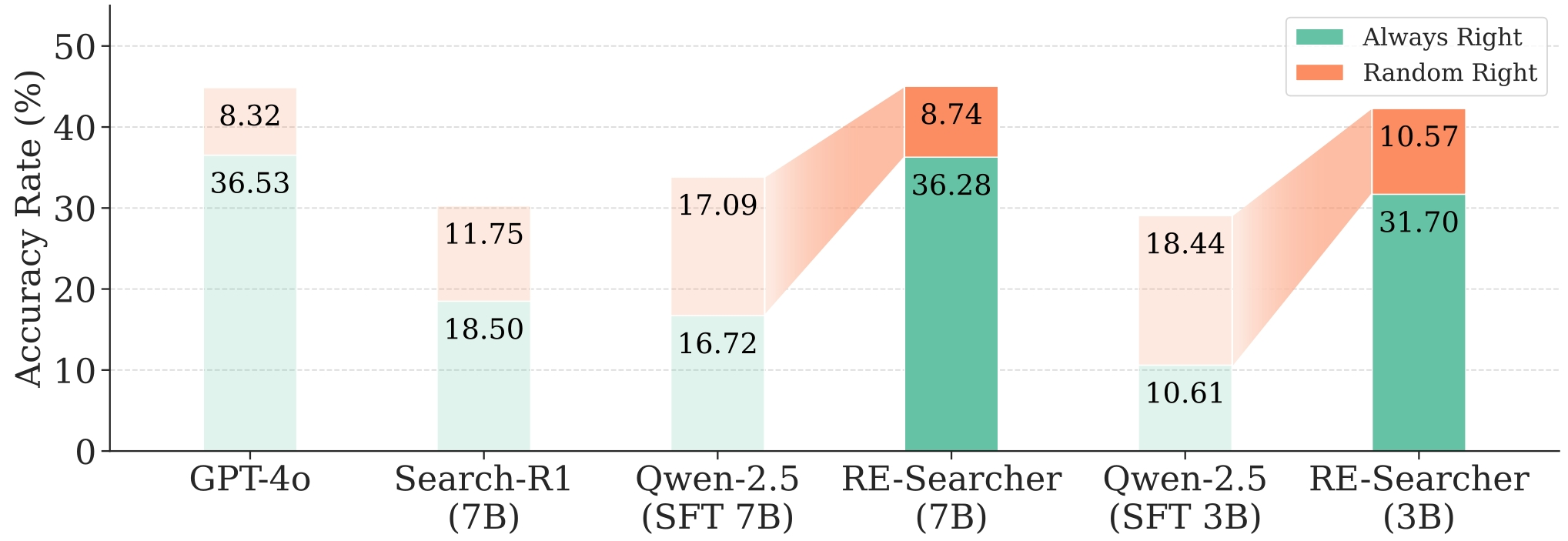}
    \caption{Analysis on the negative impacts of search fragility. The self-reflection mechanism can effectively alleviate the negative impacts of search fragility.}
    \label{fig:fragility}
\end{figure}

\subsection{Negative impacts of search fragility} 
\label{exp:nim}
We further demonstrate that the self-reflection mechanism can effectively alleviate the negative impacts of search fragility. \cref{fig:fragility} presents the Pass@k (k=2) results for GPT-4o, Search-R1, Qwen-2.5-3B-SFT, Qwen-2.5-7B-SFT, and our {\method} with both Qwen-2.5-3B-instruct and Qwen-2.5-7B-instruct as base model. In this context, the ``always right" refers to the proportion of instances where all k attempts yield the correct answer, while the ``random right" indicates the proportion of instances where at least one out of k attempts is correct.
The results clearly showcase that through training with self-reflection, the random right ratio is substantially reduced, particularly against Qwen-2.5-7B-SFT, where it decreased by approximately 8.35\%, and even more significantly against Search-R1, with a reduction of up to 3.01\%.
A surprising finding is that the random right ratio of our {\method} (7B) is 8.74\%, remarkably close to GPT-4o's 8.32\%. This proximity strongly demonstrates the effectiveness of our self-reflection mechanism in alleviating search fragility.

\begin{wrapfigure}{r}{0.52\textwidth}
    \centering
    \vspace{-1.3em}
    \includegraphics[width=\linewidth]{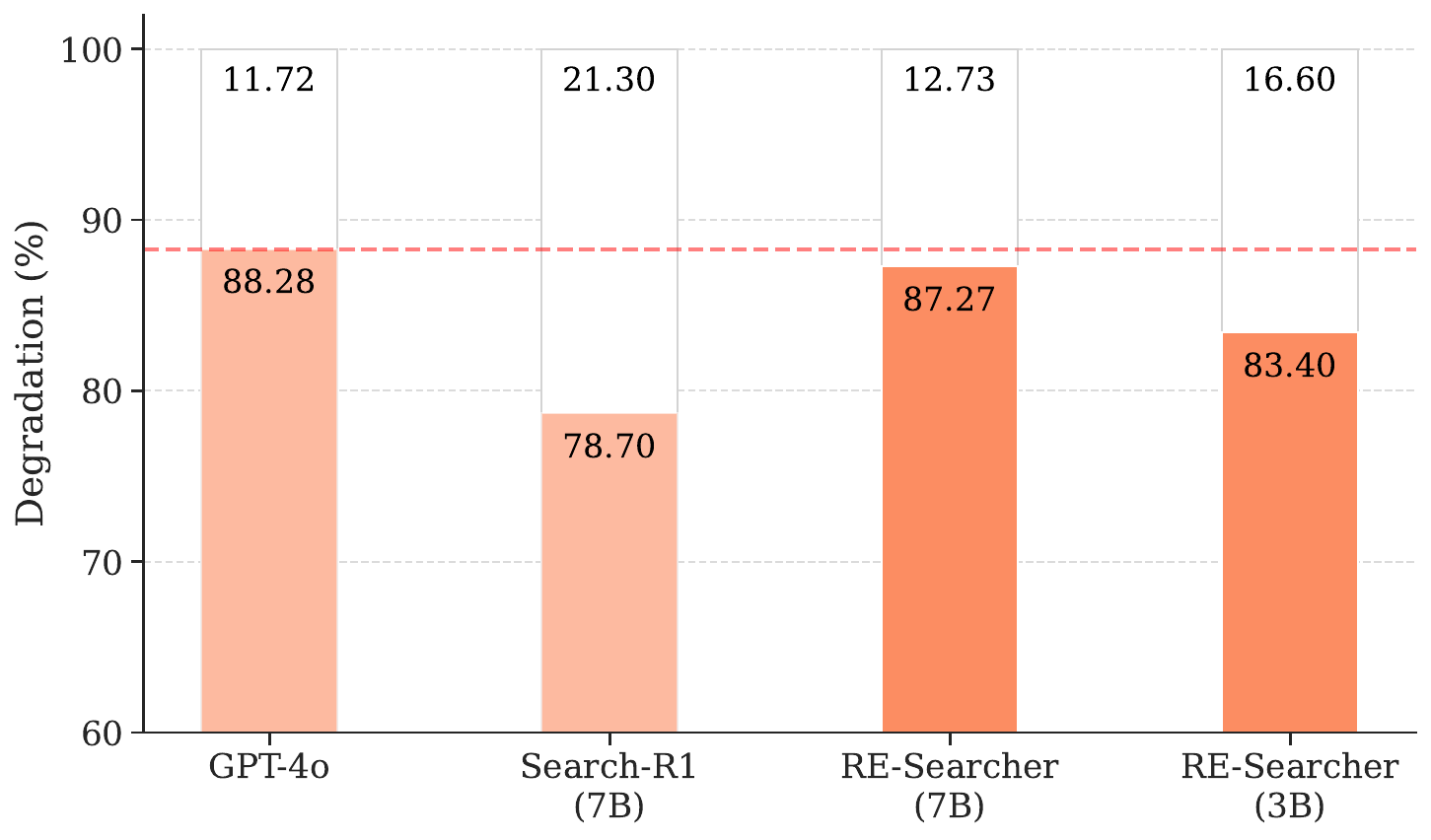}
    \vspace{-1.5em}
    \caption{Robustness analysis against disturbances. Our {\method} exhibits a substantially lower degradation.}
    \label{fig:perturbation}
\end{wrapfigure}

\subsection{Robustness against external disturbances} 
\label{exp:disturb}
Finally, we demonstrate that our proposed framework significantly enhances the model's robustness against external disturbances. To simulate real-world noise, we intentionally introduce disturbances to the queries during the first round of the search process. This is designed to both misdirect the initial search direction and challenge the model's corrective capabilities.
Specifically, we randomly employ one of the following three types of disturbances:
i) Randomly reducing a word: A word is randomly removed from the query.
ii) Randomly adding a word: A random word is inserted into the query.
iii) Randomly replacing a word with similar semantics: A word is replaced by another with a similar meaning.
All these disturbance operations are implemented by prompting GPT-4o-mini.
We then compare the proportion of instances that transition from correct to incorrect after noise injection, effectively measuring the degradation caused by disturbances. The results, presented in~\cref{fig:perturbation}, show that our {\method} exhibits a substantially lower degradation compared to Search-R1. Specifically, our framework achieves an improvement of -8.57\% in degradation relative to the Search-R1 with the same size base model. Furthermore, even our 3B model outperforms the Search-R1 (7B) in terms of robustness. Notably, our {\method} (7B) achieves a comparable degradation to GPT-4o, further underscoring the superior ability of our self-reflection mechanism to improve robustness against external disturbances.

%% file: tables/ab1.tex
\begin{table}
    \small
    \definecolor{mycolor}{rgb}{0.64, 0.87, 0.93}
    \setlength{\tabcolsep}{1.5pt}
    \renewcommand{\arraystretch}{1.2}
    \caption{Ablation on reflection reward on multi-hop datasets. The validation samples are selected with the protocol of~\cite{zheng2025deepresearcher}.}
    \vspace{5pt}
    \label{tab:ab1}
    \centering
    \scalebox{0.78}{
    \begin{tabular}{c|ccccccccc}
\toprule
 \multirow{2}{*}{\textbf{Variants}} & \multicolumn{2}{c}{\textbf{HotpotQA}} & \multicolumn{2}{c}{\textbf{2wiki}} & \multicolumn{2}{c}{\textbf{Musique}} & \multicolumn{2}{c}{\textbf{Bamboogle}} \\
\cmidrule{2-9}
& EM & F1 & EM & F1 & EM & F1 & EM & F1 \\
\midrule
 w/o reflection reward & 0.420 & 0.545 & 0.414 & 0.487 & 0.183 & 0.270 & 0.411 & 0.533
\\
\rowcolor{mycolor!25} w/ reflection reward & 0.431 (\textbf{+0.011}) & 0.544 (\textcolor{gray}{\textbf{-0.001}}) & 0.476 (\textbf{+0.062}) & 0.549 (\textbf{+0.062}) & 0.197 (\textbf{+0.014}) & 0.290 (\textbf{+0.020}) & 0.480 (\textbf{+0.069}) & 0.578 (\textbf{+0.045})
\\
\bottomrule
    \end{tabular}}
\end{table}

%% file: tables/ab2.tex
\begin{table}
    \small
    \definecolor{mycolor}{rgb}{0.64, 0.87, 0.93}
    \setlength{\tabcolsep}{4pt}
    \renewcommand{\arraystretch}{1.2}
    \caption{Ablation on reward components on in-domain and out-of-domain datasets. The validation samples are selected with the protocol of~\cite{zheng2025deepresearcher}.}
    \vspace{5pt}
    \label{tab:ab2}
    \centering
    \scalebox{0.92}{
    \begin{tabular}{l|ccccccccc}
\toprule
{\textbf{Variants}} & {\textbf{In domain}} & \textbf{Out of domain} & \textbf{AVG.} \\
\midrule
\rowcolor{mycolor!25} baseline & 0.403 & 0.395 & 0.397
\\
w/o format reward & 0.397 (\textbf{-0.006}) & 0.388 (\textbf{-0.007}) & 0.390 (\textbf{-0.007})
\\
w/o reflection reward & 0.396 (\textbf{-0.007}) & 0.387 (\textbf{-0.008}) & 0.389 (\textbf{-0.008})
\\
\bottomrule
    \end{tabular}}
\end{table}

%% file: sections/6_conclusion.tex
\section{Discussion and Conclusion}

In this paper, we investigate the instability of search agents during search and problem-solving. We identify a critical issue: complex external environments can amplify small initial errors into large deviations in the final output. To address this, we propose \method, a novel search agent that integrates goal setting with outcome reflection to counteract the fragility of search processes in complex environments. Through extensive numerical and perturbation experiments, we demonstrate that our approach substantially improves the robustness of search agents.
Nevertheless, we acknowledge that this work represents an initial step. The proposed training methodology is relatively simple, and there is considerable scope for enhancement. Future improvements could involve refining the training data, advancing the learning algorithms, and designing more sophisticated supervision signals. We believe that with these enhancements, the agent's performance in complex environments can be further elevated.

Looking ahead, the rapid progress of LLM-powered agents is enabling them to operate across an ever-wider array of external environments, i.e., often more complex and dynamic than before. While we embrace the convenience and capabilities that greater agent autonomy brings, we must also pay close attention to the complex and potentially unintended consequences of their interactions with the environment. Our future work will delve deeper into these potential issues, aiming to foster the sustainable and responsible advancement of autonomous agents.

%% file: sections/appendix.tex
\appendix
\section{Appendix}

\subsection{Experiments Details for Preliminary Analysis}
\label{sec:preliminary_exp}

\subsubsection{Stochasticity Analysis}

To investigate the instability of search agents during the search process, we constructed agents based on three distinct models: GPT-4o, Qwen2.5 3B, and Qwen2.5 7B. To ensure that the Qwen2.5 models produced outputs in the required format, we fine-tuned them using the warm-up data detailed in~\cref{sec:chat_template}. Our evaluation was conducted on a dataset of 3,197 instances selected by \cite{zheng2025deepresearcher}, with Exact Match (EM) serving as the primary metric for accuracy. Each agent was run $k=2$ times on the dataset. We categorize the outcomes as follows: questions answered correctly in all trials are labeled ``always right," while those answered correctly in some but not all trials are labeled ``random right." We then calculated the proportions of ``always right" ($P_{\text{AR}}$) and ``random right"($P_{\text{RR}}$) questions by dividing their respective counts by the total number of questions:

\begin{align}
    P_{\text{AR}} & = \frac{1}{N} \sum_{i=1}^{N} \mathbf{1}\Big\{\sum_{r=1}^{k} c_i^{(r)} = k \Big\} \\
    P_{\text{RR}} & = \frac{1}{N} \sum_{i=1}^{N} \mathbf{1}\Big\{ 1 \le \sum_{r=1}^{k} c_i^{(r)} \le k-1 \Big\}
\end{align}

Where N is the total number of the instances. $c_i^{(r)}$ is an indicator variable representing whether the answer is correct for sample $i$ in trial $r$, where a correct answer is recorded as 1 and an incorrect answer is recorded as 0.

\subsubsection{Fragility Analysis}

To quantify the impact of minor variations in search queries on the search results, we introduce three types of single-word perturbations to the keywords within the model's search trajectory: \textbf{synonym substitution}, \textbf{keyword addition}, and \textbf{keyword deletion}. We use the search engine from \cite{jin2025search} to retrieve results for both the original and the perturbed queries, yielding search result $R$ and $R'$, respectively.
Subsequently, we employ the all-MiniLM-L6-v2 model\footnote{https://huggingface.co/sentence-transformers/all-MiniLM-L6-v2} to encode each set of search results into a dense vector representation. The similarity between the original and perturbed results is then measured by computing the cosine similarity of their corresponding vectors. The formula for calculating this search result similarity is as follows:

\begin{equation}
    S(R, R') = \cos(\theta) = \frac{\vec{v} \cdot \vec{v}'}{\|\vec{v}\| \|\vec{v}'\|}
\end{equation}

where $S(R, R')$ represents the final similarity score between the original search results $R$ and the perturbed search results $R'$. $\vec{v}$ and $\vec{v}'$ represents the vector embedding of the original search results $R$ and perturbed search results $R'$ respectively.

%% file: preprint.bbl
\begin{thebibliography}{10}

\bibitem{arslan2024survey}
Muhammad Arslan, Hussam Ghanem, Saba Munawar, and Christophe Cruz.
\newblock A survey on rag with llms.
\newblock {\em Procedia computer science}, 246:3781--3790, 2024.

\bibitem{cheng2024dated}
Jeffrey Cheng, Marc Marone, Orion Weller, Dawn Lawrie, Daniel Khashabi, and Benjamin Van~Durme.
\newblock Dated data: Tracing knowledge cutoffs in large language models.
\newblock {\em arXiv preprint arXiv:2403.12958}, 2024.

\bibitem{chung2024scaling}
Hyung~Won Chung, Le~Hou, Shayne Longpre, Barret Zoph, Yi~Tay, William Fedus, Yunxuan Li, Xuezhi Wang, Mostafa Dehghani, Siddhartha Brahma, et~al.
\newblock Scaling instruction-finetuned language models.
\newblock {\em Journal of Machine Learning Research}, 25(70):1--53, 2024.

\bibitem{edge2024local}
Darren Edge, Ha~Trinh, Newman Cheng, Joshua Bradley, Alex Chao, Apurva Mody, Steven Truitt, Dasha Metropolitansky, Robert~Osazuwa Ness, and Jonathan Larson.
\newblock From local to global: A graph rag approach to query-focused summarization.
\newblock {\em arXiv preprint arXiv:2404.16130}, 2024.

\bibitem{feng2025airrag}
Wenfeng Feng, Chuzhan Hao, Yuewei Zhang, Jingyi Song, and Hao Wang.
\newblock Airrag: Activating intrinsic reasoning for retrieval augmented generation using tree-based search.
\newblock {\em arXiv preprint arXiv:2501.10053}, 2025.

\bibitem{guo2025deepseek}
Daya Guo, Dejian Yang, Haowei Zhang, Junxiao Song, Ruoyu Zhang, Runxin Xu, Qihao Zhu, Shirong Ma, Peiyi Wang, Xiao Bi, et~al.
\newblock Deepseek-r1: Incentivizing reasoning capability in llms via reinforcement learning.
\newblock {\em arXiv preprint arXiv:2501.12948}, 2025.

\bibitem{guo2024lightrag}
Zirui Guo, Lianghao Xia, Yanhua Yu, Tu~Ao, and Chao Huang.
\newblock Lightrag: Simple and fast retrieval-augmented generation.
\newblock {\em arXiv preprint arXiv:2410.05779}, 2024.

\bibitem{hao2025dynasearcher}
Chuzhan Hao, Wenfeng Feng, Yuewei Zhang, and Hao Wang.
\newblock Dynasearcher: Dynamic knowledge graph augmented search agent via multi-reward reinforcement learning.
\newblock {\em arXiv preprint arXiv:2507.17365}, 2025.

\bibitem{ho2020constructing}
Xanh Ho, Anh-Khoa~Duong Nguyen, Saku Sugawara, and Akiko Aizawa.
\newblock Constructing a multi-hop qa dataset for comprehensive evaluation of reasoning steps.
\newblock {\em arXiv preprint arXiv:2011.01060}, 2020.

\bibitem{huang2025survey}
Lei Huang, Weijiang Yu, Weitao Ma, Weihong Zhong, Zhangyin Feng, Haotian Wang, Qianglong Chen, Weihua Peng, Xiaocheng Feng, Bing Qin, et~al.
\newblock A survey on hallucination in large language models: Principles, taxonomy, challenges, and open questions.
\newblock {\em ACM Transactions on Information Systems}, 43(2):1--55, 2025.

\bibitem{hurst2024gpt}
Aaron Hurst, Adam Lerer, Adam~P Goucher, Adam Perelman, Aditya Ramesh, Aidan Clark, AJ~Ostrow, Akila Welihinda, Alan Hayes, Alec Radford, et~al.
\newblock Gpt-4o system card.
\newblock {\em arXiv preprint arXiv:2410.21276}, 2024.

\bibitem{ji2023towards}
Ziwei Ji, Tiezheng Yu, Yan Xu, Nayeon Lee, Etsuko Ishii, and Pascale Fung.
\newblock Towards mitigating llm hallucination via self reflection.
\newblock In {\em Findings of the Association for Computational Linguistics: EMNLP 2023}, pages 1827--1843, 2023.

\bibitem{jin2025search}
Bowen Jin, Hansi Zeng, Zhenrui Yue, Jinsung Yoon, Sercan Arik, Dong Wang, Hamed Zamani, and Jiawei Han.
\newblock Search-r1: Training llms to reason and leverage search engines with reinforcement learning.
\newblock {\em arXiv preprint arXiv:2503.09516}, 2025.

\bibitem{joshi2017triviaqa}
Mandar Joshi, Eunsol Choi, Daniel~S Weld, and Luke Zettlemoyer.
\newblock Triviaqa: A large scale distantly supervised challenge dataset for reading comprehension.
\newblock {\em arXiv preprint arXiv:1705.03551}, 2017.

\bibitem{kwiatkowski2019natural}
Tom Kwiatkowski, Jennimaria Palomaki, Olivia Redfield, Michael Collins, Ankur Parikh, Chris Alberti, Danielle Epstein, Illia Polosukhin, Jacob Devlin, Kenton Lee, et~al.
\newblock Natural questions: a benchmark for question answering research.
\newblock {\em Transactions of the Association for Computational Linguistics}, 7:453--466, 2019.

\bibitem{lewis2020retrieval}
Patrick Lewis, Ethan Perez, Aleksandra Piktus, Fabio Petroni, Vladimir Karpukhin, Naman Goyal, Heinrich K{\"u}ttler, Mike Lewis, Wen-tau Yih, Tim Rockt{\"a}schel, et~al.
\newblock Retrieval-augmented generation for knowledge-intensive nlp tasks.
\newblock {\em Advances in neural information processing systems}, 33:9459--9474, 2020.

\bibitem{li2025codei}
Junlong Li, Daya Guo, Dejian Yang, Runxin Xu, Yu~Wu, and Junxian He.
\newblock Codei/o: Condensing reasoning patterns via code input-output prediction.
\newblock {\em arXiv preprint arXiv:2502.07316}, 2025.

\bibitem{li2025search}
Xiaoxi Li, Guanting Dong, Jiajie Jin, Yuyao Zhang, Yujia Zhou, Yutao Zhu, Peitian Zhang, and Zhicheng Dou.
\newblock Search-o1: Agentic search-enhanced large reasoning models.
\newblock {\em arXiv preprint arXiv:2501.05366}, 2025.

\bibitem{ma2023query}
Xinbei Ma, Yeyun Gong, Pengcheng He, Hai Zhao, and Nan Duan.
\newblock Query rewriting in retrieval-augmented large language models.
\newblock In {\em Proceedings of the 2023 Conference on Empirical Methods in Natural Language Processing}, pages 5303--5315, 2023.

\bibitem{mallen2022not}
Alex Mallen, Akari Asai, Victor Zhong, Rajarshi Das, Daniel Khashabi, and Hannaneh Hajishirzi.
\newblock When not to trust language models: Investigating effectiveness of parametric and non-parametric memories.
\newblock {\em arXiv preprint arXiv:2212.10511}, 2022.

\bibitem{mei20252}
Jianbiao Mei, Tao Hu, Daocheng Fu, Licheng Wen, Xuemeng Yang, Rong Wu, Pinlong Cai, Xinyu Cai, Xing Gao, Yu~Yang, et~al.
\newblock O2-searcher: A searching-based agent model for open-domain open-ended question answering.
\newblock {\em arXiv preprint arXiv:2505.16582}, 2025.

\bibitem{minaee2024large}
Shervin Minaee, Tomas Mikolov, Narjes Nikzad, Meysam Chenaghlu, Richard Socher, Xavier Amatriain, and Jianfeng Gao.
\newblock Large language models: A survey.
\newblock {\em arXiv preprint arXiv:2402.06196}, 2024.

\bibitem{OpenAI_DeepResearch}
{OpenAI}.
\newblock Introducing deep research.
\newblock \url{https://openai.com/zh-Hans-CN/index/introducing-deep-research/}, February 2025.
\newblock Accessed: 2025-09-23.

\bibitem{press2022measuring}
Ofir Press, Muru Zhang, Sewon Min, Ludwig Schmidt, Noah~A Smith, and Mike Lewis.
\newblock Measuring and narrowing the compositionality gap in language models.
\newblock {\em arXiv preprint arXiv:2210.03350}, 2022.

\bibitem{qwen2025qwen25technicalreport}
Qwen, :, An~Yang, Baosong Yang, Beichen Zhang, Binyuan Hui, Bo~Zheng, Bowen Yu, Chengyuan Li, Dayiheng Liu, Fei Huang, Haoran Wei, Huan Lin, Jian Yang, Jianhong Tu, Jianwei Zhang, Jianxin Yang, Jiaxi Yang, Jingren Zhou, Junyang Lin, Kai Dang, Keming Lu, Keqin Bao, Kexin Yang, Le~Yu, Mei Li, Mingfeng Xue, Pei Zhang, Qin Zhu, Rui Men, Runji Lin, Tianhao Li, Tianyi Tang, Tingyu Xia, Xingzhang Ren, Xuancheng Ren, Yang Fan, Yang Su, Yichang Zhang, Yu~Wan, Yuqiong Liu, Zeyu Cui, Zhenru Zhang, and Zihan Qiu.
\newblock Qwen2.5 technical report, 2025.

\bibitem{schick2023toolformer}
Timo Schick, Jane Dwivedi-Yu, Roberto Dess{\`\i}, Roberta Raileanu, Maria Lomeli, Eric Hambro, Luke Zettlemoyer, Nicola Cancedda, and Thomas Scialom.
\newblock Toolformer: Language models can teach themselves to use tools.
\newblock {\em Advances in Neural Information Processing Systems}, 36:68539--68551, 2023.

\bibitem{shah2025beyond}
Agam Shah, Liqin Ye, Sebastian Jaskowski, Wei Xu, and Sudheer Chava.
\newblock Beyond the reported cutoff: Where large language models fall short on financial knowledge.
\newblock {\em arXiv preprint arXiv:2504.00042}, 2025.

\bibitem{shao2024deepseekmath}
Zhihong Shao, Peiyi Wang, Qihao Zhu, Runxin Xu, Junxiao Song, Xiao Bi, Haowei Zhang, Mingchuan Zhang, YK~Li, Y~Wu, et~al.
\newblock Deepseekmath: Pushing the limits of mathematical reasoning in open language models.
\newblock {\em arXiv preprint arXiv:2402.03300}, 2024.

\bibitem{song2025r1}
Huatong Song, Jinhao Jiang, Wenqing Tian, Zhipeng Chen, Yuhuan Wu, Jiahao Zhao, Yingqian Min, Wayne~Xin Zhao, Lei Fang, and Ji-Rong Wen.
\newblock R1-searcher++: Incentivizing the dynamic knowledge acquisition of llms via reinforcement learning.
\newblock {\em arXiv preprint arXiv:2505.17005}, 2025.

\bibitem{sun2025zerosearch}
Hao Sun, Zile Qiao, Jiayan Guo, Xuanbo Fan, Yingyan Hou, Yong Jiang, Pengjun Xie, Yan Zhang, Fei Huang, and Jingren Zhou.
\newblock Zerosearch: Incentivize the search capability of llms without searching.
\newblock {\em arXiv preprint arXiv:2505.04588}, 2025.

\bibitem{tonmoy2024comprehensive}
SMTI Tonmoy, SM~Zaman, Vinija Jain, Anku Rani, Vipula Rawte, Aman Chadha, and Amitava Das.
\newblock A comprehensive survey of hallucination mitigation techniques in large language models.
\newblock {\em arXiv preprint arXiv:2401.01313}, 6, 2024.

\bibitem{trivedi2022interleaving}
Harsh Trivedi, Niranjan Balasubramanian, Tushar Khot, and Ashish Sabharwal.
\newblock Interleaving retrieval with chain-of-thought reasoning for knowledge-intensive multi-step questions.
\newblock {\em arXiv preprint arXiv:2212.10509}, 2022.

\bibitem{trivedi2022musique}
Harsh Trivedi, Niranjan Balasubramanian, Tushar Khot, and Ashish Sabharwal.
\newblock Musique: Multihop questions via single-hop question composition.
\newblock {\em Transactions of the Association for Computational Linguistics}, 10:539--554, 2022.

\bibitem{wang2025otc}
Hongru Wang, Cheng Qian, Wanjun Zhong, Xiusi Chen, Jiahao Qiu, Shijue Huang, Bowen Jin, Mengdi Wang, Kam-Fai Wong, and Heng Ji.
\newblock Otc: Optimal tool calls via reinforcement learning.
\newblock {\em arXiv e-prints}, pages arXiv--2504, 2025.

\bibitem{wang2025stepsearch}
Ziliang Wang, Xuhui Zheng, Kang An, Cijun Ouyang, Jialu Cai, Yuhang Wang, and Yichao Wu.
\newblock Stepsearch: Igniting llms search ability via step-wise proximal policy optimization.
\newblock {\em arXiv preprint arXiv:2505.15107}, 2025.

\bibitem{wei2022chain}
Jason Wei, Xuezhi Wang, Dale Schuurmans, Maarten Bosma, Fei Xia, Ed~Chi, Quoc~V Le, Denny Zhou, et~al.
\newblock Chain-of-thought prompting elicits reasoning in large language models.
\newblock {\em Advances in neural information processing systems}, 35:24824--24837, 2022.

\bibitem{wu2025kg}
Rong Wu, Pinlong Cai, Jianbiao Mei, Licheng Wen, Tao Hu, Xuemeng Yang, Daocheng Fu, and Botian Shi.
\newblock Kg-traces: Enhancing large language models with knowledge graph-constrained trajectory reasoning and attribution supervision.
\newblock {\em arXiv preprint arXiv:2506.00783}, 2025.

\bibitem{yang2024qwen2}
An~Yang, Baosong Yang, Beichen Zhang, Binyuan Hui, Bo~Zheng, Bowen Yu, Chengyuan Li, Dayiheng Liu, Fei Huang, Haoran Wei, et~al.
\newblock Qwen2. 5 technical report.
\newblock {\em arXiv preprint arXiv:2412.15115}, 2024.

\bibitem{yang2018hotpotqa}
Zhilin Yang, Peng Qi, Saizheng Zhang, Yoshua Bengio, William~W Cohen, Ruslan Salakhutdinov, and Christopher~D Manning.
\newblock Hotpotqa: A dataset for diverse, explainable multi-hop question answering.
\newblock {\em arXiv preprint arXiv:1809.09600}, 2018.

\bibitem{yao2023react}
Shunyu Yao, Jeffrey Zhao, Dian Yu, Nan Du, Izhak Shafran, Karthik Narasimhan, and Yuan Cao.
\newblock React: Synergizing reasoning and acting in language models.
\newblock In {\em International Conference on Learning Representations (ICLR)}, 2023.

\bibitem{yu2024rankrag}
Yue Yu, Wei Ping, Zihan Liu, Boxin Wang, Jiaxuan You, Chao Zhang, Mohammad Shoeybi, and Bryan Catanzaro.
\newblock Rankrag: Unifying context ranking with retrieval-augmented generation in llms.
\newblock {\em Advances in Neural Information Processing Systems}, 37:121156--121184, 2024.

\bibitem{zheng2025deepresearcher}
Yuxiang Zheng, Dayuan Fu, Xiangkun Hu, Xiaojie Cai, Lyumanshan Ye, Pengrui Lu, and Pengfei Liu.
\newblock Deepresearcher: Scaling deep research via reinforcement learning in real-world environments.
\newblock {\em arXiv preprint arXiv:2504.03160}, 2025.

\end{thebibliography}
